\documentclass[11pt]{article}

\usepackage[preprint]{acl}
\usepackage{times}
\usepackage{latexsym}
\usepackage[T1]{fontenc}
\usepackage[utf8]{inputenc}
\usepackage{microtype}
\usepackage{inconsolata}
\usepackage{graphicx}
\usepackage{booktabs}
\usepackage{amsmath}
\usepackage{multirow}
\usepackage{xcolor}
\usepackage{tabularx}
\usepackage{enumitem}

\newcommand{\ci}[2]{\,{\scriptsize[#1,\,#2]}}
\newcommand{\qnone}{\textnormal{\textsc{none}}}
\newcommand{\cno}{\textit{no}}
\newcommand{\cyes}{\textit{yes}}
\newcommand{\cpart}{\textit{partial}}
\title{Better Retrievers, Worse Labels: How Incomplete Relevance Judgments\\
Distort Abstention Evaluation in Conversational RAG}

\author{Ivan Kochergin \\
  Letovo School, Moscow, Russia \\
  \texttt{2027kochergin.is@student.letovo.ru} \\\And
  Vasily Konovalov \\
  AXXX \\
  \texttt{vaskonov@yahoo.com}}

\begin{document}
\maketitle

\begin{abstract}
Conversational RAG benchmarks such as MTRAG and RETECO evaluate retrieval and generation end to end: each system answers from its own retrieval, so abstention must be scored against whether each retrieved context is sufficient. A convenient shortcut reads sufficiency off the relevance judgments and treats a context without gold documents as unanswerable. We re-run the official baselines exactly and test this shortcut on RECOR (RETECO track~2) with four retrievers (nDCG@10 from .18 to .59) and three Qwen3 readers. Two open LLM judges and a panel of four small open models label each context by content, and a model of another family audits 250 contexts blind, with adjudication. The stronger the retriever, the larger the share of no-gold contexts that still contain the answer (28\% to 72\%) and the lower the shortcut's precision (.68 to .10; .70 to .21 under the audit). A reader agrees: Qwen3-4B answers 62--67\% of these false misses correctly, against 23--30\% of no-gold contexts judged insufficient and 1\% with no documents. In most false misses, the answer comes from a hard negative mined for the same question, another turn's gold or a neighbouring chunk of the gold page. On MTRAG, whose judgments come from annotators, precision drops from .72 to .50. Under the shortcut, answerability AUROC is up to .19 too low, and the best sufficiency detector appears to be a retrieval score that never reads the documents. We release code, runs and labels, and recommend scoring refusals against content labels, not the presence of gold.
\end{abstract}

\section{Introduction}

A retrieval-augmented reader should answer when the retrieved documents support an answer and abstain when they do not~\cite{joren2025sufficient,song2025trustscore}. In conversational search this decision is frequent, because turns are elliptical and first-stage retrieval often misses. Recent benchmarks therefore evaluate retrieval and generation end to end, with each system answering from what it retrieved: MTRAG~\cite{katsis2025mtrag}, MTRAGEval at SemEval-2026~\cite{rosenthal2026semeval} and RETECO at SemEval-2027~\cite{abdallah2026reteco}. RETECO pairs two settings on the same turns: answering from the gold documents (task 2b) and from one's own retrieval (task 2c).

Scoring abstention end to end requires knowing, for each retrieved context, whether it suffices. Labelling every context of every system by hand is infeasible, so the convenient protocol reads sufficiency off the relevance judgments: a context that contains none of the turn's gold documents is \emph{insufficient}. At the question level, MTRAG and MTRAGEval call a turn unanswerable when annotators found no relevant passage~\cite{katsis2025mtrag,rosenthal2026semeval}, and unanswerability is created by deleting gold passages from the index~\cite{chandradevan2026search}. At the context level, insufficiency detectors are trained and tested on whether the annotated passages are present in the retrieved set~\cite{chen2026rinse}. The 2b/2c pairing invites the same reading: 2b is sufficient by construction, and a 2c context without gold looks like natural unanswerability.

\begin{figure*}[t]
\centering\small
\begin{tabularx}{\textwidth}{@{}Xcc@{}}
\toprule
\multicolumn{3}{@{}p{\textwidth}@{}}{\textbf{Question:} \emph{In glucose meters, how come they only pick up glucose and not other sugars like sucrose in water?}}\\
\addlinespace[2pt]
\multicolumn{3}{@{}p{\textwidth}@{}}{\textbf{Retrieved document 2} (qrels: negative): ``\ldots{} based upon the enzyme glucose oxidase. Since an enzyme is used, it is very specific for only D-glucose \ldots'' The gold document contains this passage verbatim but was not retrieved.}\\
\midrule
 & qrels label & content label\\
\cmidrule(l){2-3}
Context & insufficient & sufficient\\
Reader abstains & correct & wrong\\
Reader answers correctly & wrong & correct\\
\bottomrule
\end{tabularx}
\caption{A false miss of the Fusion retriever (medical sciences). The retrieved document contains the answer but is a negative in the qrels, so the two labels grade the same reader behaviour in opposite ways.}
\label{fig:teaser}
\end{figure*}

Relevance judgments are incomplete, and stronger systems retrieve more unjudged relevant documents~\cite{arabzadeh2022shallow,weller2025rank1}. Fig.~\ref{fig:teaser} shows the consequence: the retriever finds the answer in a document the qrels do not credit, and the protocol rewards the reader that refuses. No earlier use measures how its validity changes with the retriever. In this reproducibility paper we re-run the official RETECO baselines exactly, reproduce the context-level label on RECOR~\cite{ali2026recor} (different team, different setup), and generalise it along what earlier uses held fixed: the retriever, the source of the label and the benchmark. We ask:

\begin{enumerate}[label=\textbf{RQ\arabic*},widest=4,leftmargin=*,align=left,labelsep=0.5em]
\item Does the qrels-derived sufficiency label stay valid as the retriever improves?
\item Are the content-based labels that contradict it trustworthy?
\item Does the choice of label change conclusions about abstention?
\item Does the effect generalise to MTRAG, whose qrels come from annotators and whose corpus has no mined negatives?
\end{enumerate}

The protocol's precision falls from .68 to .10 as nDCG@10 rises from .18 to .59 (\S\ref{sec:rq1}). Independent checks support the content labels: a judge-free count of gold copies, a blind audit by a model of another family, a negative control, human judgments on MTRAG and a reader's own answers; four small open judges reproduce the decline but are more lenient (\S\ref{sec:rq2}). The label changes which readers and detectors look good (\S\ref{sec:rq3}), and the decline replicates on MTRAG (\S\ref{sec:rq4}). We argue that the protocol can reverse conclusions whenever retrievers of different strength are compared, and that abstention in end-to-end benchmarks should be scored against content-based labels of each retrieved context (\S\ref{sec:discussion}). Code, runs, model outputs, labels and audit annotations are available.\footnote{\url{https://github.com/iskochergin/reteco-2026}}

\section{Related Work}

\paragraph{Sufficiency and abstention in RAG.}
Abstention has been studied for reading comprehension with unanswerable questions~\cite{rajpurkar2018know,trivedi2022musique}, for LLMs in general~\cite{wen2025know,kirichenko2025abstentionbench}, and for RAG, where the reader must notice that the retrieved text does not support an answer~\cite{chen2024benchmarking,peng2025unanswerability}. \citet{joren2025sufficient} define \emph{sufficient context} by content and label it with an LLM autorater; \citet{song2025trustscore} score groundedness and refusal per context against gold claims. Refusal after a retrieval miss has been tabulated per retriever~\cite{vondaniken2026rat}. None of these compares content labels with qrels-derived labels across retrievers.

\paragraph{Unanswerability in shared tasks.}
The MTRAG and SemEval-2026 Task~8 organizers acknowledge likely missed relevant passages and a likely bias toward the ELSER retriever used to build the data~\cite{katsis2025mtrag,rosenthal2026mtragun}, and the winning retrieval system attributes part of ELSER's advantage to this bias~\cite{athanasiou2026ails}. The task overview notes one unanswerable task for which 27 systems retrieved relevant passages~\cite{rosenthal2026semeval}. Participants report that all unanswerable generation items come with empty reference passages, a shortcut for abstention~\cite{bondarenko2026raguteam}, and that IDK conditioning penalises hedged grounded answers~\cite{dutra2026nlpceia}. Outside MTRAG, CLAPNQ, whose corpus is one MTRAG domain, attributes its drop in unanswerable accuracy under retrieval to questions that become answerable from the retrieved passages~\cite{rosenthal2025clapnq}; in TyDi~QA, retrieval misses account for 50--74\% of the annotated unanswerable questions~\cite{asai2021challenges}; CRAG scores a missing answer as zero whether or not the retrieved pages contained it~\cite{yang2024crag}; and LiveRAG notes that reference answers built from one or two documents can penalise correct answers found elsewhere~\cite{carmel2025liverag}. These observations concern single items or ranking metrics; none measures how often a context without gold contains the answer as retrievers improve.

\paragraph{Incomplete relevance judgments.}
Pooling leaves relevant documents unjudged and biases evaluation against systems that retrieve them~\cite{zobel1998reliable,buckley2004retrieval,yilmaz2006estimating}. Shallow judgments miss many relevant passages in MS~MARCO and BEIR, where dense retrievers suffer most~\cite{arabzadeh2022shallow,thakur2021beir}, and mined hard negatives are often relevant~\cite{qu2021rocketqa,moreira2024nvretriever,thakur2025hard}; on TREC DL19, the share of unjudged top-10 documents that are relevant rises from 50\% for MonoT5 to 86--88\% for reasoning rerankers~\cite{weller2025rank1}. BRIGHT corpora contain unlabelled copies of gold documents that penalise stronger retrievers more~\cite{sharifymoghaddam2026lighting}. LLM judgments can fill such holes~\cite{thomas2024large,upadhyay2024umbrela}, with risks of circularity~\cite{clarke2025llm}. We show that the same incompleteness becomes a retriever-dependent bias in a binary label that decides which abstentions count as correct.

\section{Reproduction Setup}
\label{sec:setup}

\paragraph{Data.}
RETECO track~2 uses RECOR~\cite{ali2026recor}. Each dialogue expands a question from StackExchange (drones, politics, law, medical sciences, hardware) or from BRIGHT~\cite{su2025bright} (economics, psychology, biology, sustainable living, robotics, earth science) into 3--5 short, context-dependent turns. Every turn has gold documents and a reference answer generated by GPT-4.1 from verified atomic facts; each domain corpus (16k--121k documents) includes negatives mined for each question. We use the 858 dev turns with gold (211 dialogues). As the task allows, the official dialogue history contains the reference answers of earlier turns.

\paragraph{The protocol.}
The \textbf{qrels label} counts the gold documents in a top-$k$ context ($k{=}5$): \textsc{all}, \textsc{part} or \qnone; the protocol treats \qnone{} as insufficient. Re-running the official starter kit reproduces the published nDCG@10 of both BM25 baselines (turn only, and turn plus the official history) in all 11 domains to four decimals (dev macro .183 and .438). We then build a ladder of first stages that the protocol has not been tested on:
\textbf{BM25-turn} searches with the current turn only;
\textbf{BM25-hist} is the official baseline;
\textbf{Fusion} min-max fuses BM25 runs over three LLM query expansions (a Query2doc passage~\cite{wang2023query2doc}, a rewrite with a passage, and a rewrite with a HyDE passage~\cite{gao2023precise});
\textbf{Hybrid} adds a dense retriever (reason-embed-qwen3-4b~\cite{chen2026reasonembed}) to Fusion.
All weights were chosen on train. Dev nDCG@10 is .183, .438, .550 and .591, and the share of turns with at least one gold document in the top-5 is .28, .65, .77 and .82.

\paragraph{Readers.}
Qwen3-1.7B, Qwen3-4B-Instruct-2507 and Qwen3-8B~\cite{yang2025qwen3} answer from the numbered documents and put an explicit status, \texttt{ANSWERABLE} or \texttt{UNANSWERABLE}, on the first line; we read $p(\textsc{answerable})$ from the first-token distribution. Each reader answers every turn from the gold documents (2b), from the top-5 of each retriever (2c), and without documents. A refusal is an \texttt{UNANSWERABLE} status; responses without a status line (at most 75 of 858 per context set) are excluded from refusal rates and AUROCs. We use vLLM~\cite{kwon2023vllm}.

\paragraph{Content labels.}
A judge reads the dialogue, the latest question, the reference answer and the five numbered documents (without IDs), and decides whether the documents support the core of the reference answer: \cyes, \cpart{} (some but not all central points) or \cno. It also names the supporting documents. Two open judges of different families label all $858\times5=4{,}290$ contexts (the top-5 of each retriever and the gold contexts): phi-4~\cite{abdin2024phi4} (primary) and Qwen3-14B~\cite{yang2025qwen3}. To test whether the decline depends on judge size or family, four small open models from four families (two of them shared with the large judges) label the same contexts with the same prompt: Qwen3-4B-Instruct-2507, Phi-4-mini-instruct (3.8B), SmolLM3-3B and Granite-3.3-8B-instruct; their \emph{panel} label is the majority, with ties counted as \cpart. Finally, two runs of Claude Opus~5.5 annotate 250 contexts blind (100 Fusion contexts, half \qnone, and 50 \qnone{} contexts from each other retriever), shuffled and without IDs, so that the annotator sees neither the retriever nor the qrels label. The two runs follow different procedures (holistic versus claim by claim) and agree on 97\% of items (self-consistency of one model, not inter-annotator agreement). A third run adjudicates. Binary metrics compare \cyes{} with \cno{} and drop \cpart.

\paragraph{Statistics.}
Confidence intervals are 95\% percentile intervals from 1{,}000 bootstrap resamples of turns (seed 0); differences are paired. Resampling dialogues instead moves interval bounds by at most .03.

\paragraph{Scope and process of the reproduction.}
Re-run exactly: the official BM25 baselines. Reproduced: the context-level qrels label as used for insufficiency detection~\cite{chen2026rinse}, a deterministic function of the qrels and the run; RETECO itself does not score abstention, so we apply the label to its 2b/2c pairing. Generalised: the first-stage retriever, the source of the sufficiency label, and the benchmark (MTRAG). Everything needed was public (starter kit, qrels, MTRAG annotations), so no contact with the original authors was required. Two difficulties arose: the kit's format checker needs an explicit document key for track~2, and the query expansions behind Fusion and the hybrid came from a closed model, Claude Sonnet~5 (regenerated with an open model, the Query2doc passages give .453 instead of .546 with BM25). LLM inference ran on single A100 or L4 GPUs in bfloat16 with greedy decoding, and dense corpus embeddings in float16 (partly on T4); prompts, model identifiers and settings are released with the artifacts, and every number in the paper is recomputed by the provided scripts from saved outputs, including the query expansions and audit annotations.

\section{Results}

\subsection{RQ1: The qrels label degrades with retrieval quality}
\label{sec:rq1}

\begin{figure*}[t]
\centering
\includegraphics[width=\textwidth]{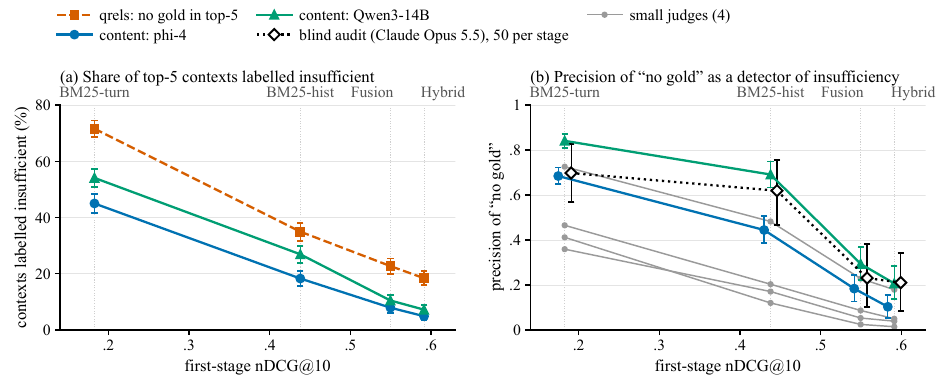}
\caption{(a) Share of top-5 contexts labelled insufficient by the qrels label (no gold) and by the two content judges (\cno; \cpart{} counted as sufficient), against first-stage nDCG@10. (b) Precision of ``no gold'' as a detector of content insufficiency under each labeller; the audit covers 50 \qnone{} contexts per retriever; grey: the four small judges of Table~\ref{tab:judges}. Bars: 95\% bootstrap CIs.}
\label{fig:bias}
\end{figure*}

With \cpart{} counted as sufficient, the qrels label overstates insufficiency, and the overstatement grows along the ladder (Fig.~\ref{fig:bias}a): 71.6\% of BM25-turn contexts lack gold and 45.0\% lack the answer according to phi-4 (1.6$\times$), while for the hybrid the figures are 18.4\% and 4.9\% (3.8$\times$). Read as a detector of insufficient context, ``no gold'' has precision .68\ci{.65}{.72} on BM25-turn, .44 on BM25-hist, .18 on Fusion and .10\ci{.05}{.16} on the hybrid (Fig.~\ref{fig:bias}b). Every consecutive step is significant, including Fusion to the hybrid ($-.08$\ci{$-$.14}{$-$.02}), and precision falls in all 11 domains. The share of \qnone{} contexts that a judge finds sufficient (\emph{false misses}) rises from 28\% to 45\%, 68\% and 72\%.

Much of the drop comes from the base rate: \cno{} contexts become rare on strong retrievers (386, 157, 68 and 42), so false misses outnumber them. But abstention evaluation depends on precision, because it sets how many sufficient contexts sit in the class against which refusals are scored. The label also becomes less informative beyond the base rate: recall minus the false-alarm rate (Youden's $J$) falls from .52 to .15 (difference $-.37$\ci{$-$.51}{$-$.22}). The overstatement grows only under this convention; precision falls under either: with \cpart{} counted as insufficient, the ratio is 1.07--1.30$\times$ under phi-4 with no trend and below 1 under Qwen3-14B on the three stronger retrievers, while precision still falls from .72 to .28 and the false-miss share, now including \cpart, rises from 39\% to 92\%.

\subsection{RQ2: Are the content labels right?}
\label{sec:rq2}

A skeptic can object that LLM judges credit on-topic documents too readily, and that this leniency, rather than missing qrels, makes strong retrievers look sufficient. We test the content labels in five ways.

\paragraph{Without any judge.}
Some false misses need no judgment: the context contains a near-verbatim copy of a gold document of the turn (at least 80\% of its word 8-grams, or character 5-gram Jaccard ${\geq}.7$), so it is relevant by the qrels' own standard and is missed only because the copy has another ID. Such contexts make up 1.6\%\ci{0.7}{2.8} of \qnone{} contexts on BM25-turn, 6.0\% on BM25-hist, 16.4\%\ci{11.2}{21.7} on Fusion and 9.5\% on the hybrid: a judge-free lower bound on how often ``no gold'' is wrong, ten times higher on Fusion than on BM25-turn.

\paragraph{Other labellers.}
Table~\ref{tab:judges} gives the precision of ``no gold'' under every labeller. Qwen3-14B marks more contexts insufficient than phi-4, but its precision falls in the same way (.84 to .21), and so does the blind audit's (.70, .62, .23, .21). Each of the four small judges, and their panel, reproduces the decline with a confidence interval that excludes zero ($-.32$ to $-.55$). The small judges are, however, more lenient: they label 7--31\% of the negative-control contexts \cyes{} (Table~\ref{tab:judges}, last column), three of them have lower precision than both large judges at every stage, and they agree with each other and with the audit only fairly (Fleiss' $\kappa$ .26 on three labels; $\kappa$ .18--.25 with the audit). They confirm the direction of the effect, not its size. Among the larger labellers agreement is fair to moderate: Cohen's $\kappa$~\cite{cohen1960coefficient} is .44 between phi-4 and Qwen3-14B on all 4{,}290 contexts, and .38 and .49 between the audit and the two judges; on the binary \cyes/\cno{} decision it is .74, .59 and .79, and raw three-label agreement is .70, .64 and .68. The large judges are not uniformly lenient: phi-4 labels 59 of the 858 gold contexts \cno{} (Qwen3-14B 56) and names as evidence only about half of the near-verbatim copies of gold that Fusion retrieves. On the stronger retrievers $\kappa$ falls, but raw agreement and Gwet's AC1~\cite{gwet2008computing} do not, consistent with the rising share of \cyes. We therefore draw conclusions from stage-level trends, on which all seven labellers agree.

\begin{table*}[t]
\centering
\caption{Precision of ``no gold'' as a detector of content insufficiency (\cno; \cpart{} excluded) under each labeller; its change from BM25-turn to the hybrid with 95\% CI; agreement ($\kappa$, three labels) with the blind audit on its 250 contexts; and the share of \cyes{} in the negative control (contexts of other dialogues). Panel: majority of the four small judges, ties as \cpart. Unparsable outputs (at most 27 of 858 per context set) are dropped from precision; control shares are over all 858 contexts.}
\label{tab:judges}
\setlength{\tabcolsep}{3.5pt}
\begin{tabular}{@{}lccccccc@{}}
\toprule
Labeller & Turn & Hist & Fusion & Hybrid & Hybrid $-$ Turn & $\kappa$ audit & control \\
\midrule
phi-4 (14B) & .68 & .44 & .18 & .10 & $-.58$\ci{$-$.64}{$-$.52} & .38 & 4.1\% \\
Qwen3-14B & .84 & .69 & .29 & .21 & $-.63$\ci{$-$.71}{$-$.56} & .49 & 0.0\% \\
\addlinespace[2pt]
Qwen3-4B-Instruct & .36 & .17 & .05 & .04 & $-.32$\ci{$-$.37}{$-$.28} & .24 & 6.8\% \\
Phi-4-mini (3.8B) & .73 & .48 & .23 & .18 & $-.55$\ci{$-$.62}{$-$.47} & .25 & 7.7\% \\
SmolLM3-3B & .47 & .20 & .09 & .05 & $-.42$\ci{$-$.47}{$-$.36} & .18 & 30.7\% \\
Granite-3.3-8B & .41 & .12 & .02 & .01 & $-.40$\ci{$-$.45}{$-$.35} & .18 & 12.4\% \\
Panel of four & .49 & .19 & .04 & .02 & $-.47$\ci{$-$.52}{$-$.42} & .25 & 6.5\% \\
\addlinespace[2pt]
Blind audit & .70 & .62 & .23 & .21 & & -- & -- \\
\bottomrule
\end{tabular}
\end{table*}

\paragraph{Negative control.}
Each turn is paired with the hybrid's top-5 for a turn of another dialogue in the same domain, with the turn's own gold removed. Qwen3-14B labels none of the 858 contexts \cyes, phi-4 labels 35 (4.1\%), and the small judges 6.8--30.7\% (Table~\ref{tab:judges}). Correcting for phi-4's leniency changes the false-miss rates little (28\% to 25\% on BM25-turn, 72\% to 70\% on the hybrid). This rules out indiscriminate leniency, but not leniency toward documents about the same question; the next two checks address that.

\paragraph{A reader answers from false-miss contexts.}
If false misses really contain the answer, a reader should answer them correctly more often than other \qnone{} contexts, and should not manage without the documents. Two LLM answer judges (phi-4 and Qwen3-8B) compare each response with the reference answer without seeing any sufficiency label; an answer counts as correct if the reader answered and correctness is at least 4 of 5. On BM25-turn, BM25-hist and Fusion (answer judges were not run on the hybrid), Qwen3-4B answers 62--67\% of false misses correctly, nearly always faithfully to the documents, against 23--30\% of true misses (\qnone{} contexts that phi-4 judges insufficient) and 1\% of the same turns without documents (paired difference $+.61$ to $+.66$, every lower bound above $+.53$). The Qwen3-8B answer judge, which never labels sufficiency, gives the same picture (Table~\ref{tab:leak}). Since Qwen3-4B abstains on 99\% of turns without documents, as instructed, this control shows that its correct answers come from the retrieved documents, not that it lacks the knowledge.

\begin{table}[t]
\centering\small
\caption{Share of correct Qwen3-4B answers (answered, correctness $\geq4$ of 5 by the phi-4 answer judge) on \qnone{} contexts that phi-4 judges sufficient (false misses) or insufficient (true misses), and on the false-miss turns without documents. $n$: false misses / true misses. With the Qwen3-8B answer judge every cell is within .04.}
\label{tab:leak}
\setlength{\tabcolsep}{5pt}
\begin{tabular}{@{}lccc@{}}
\toprule
Context & BM25-turn & BM25-hist & Fusion \\
\midrule
false miss & .67 & .62 & .65 \\
true miss & .23 & .23 & .30 \\
no documents & .01 & .01 & .01 \\
\addlinespace[2pt]
$n$ & 173 / 375 & 135 / 108 & 133 / 30 \\
\bottomrule
\end{tabular}
\end{table}

\paragraph{Human judgments on MTRAG.}
MTRAG records which passages its annotators saw and rejected for a turn. Shown one passage at a time, phi-4 labels 59\% of gold passages \cyes{} and only 8\%\ci{6}{11} of rejected passages; 1 of 109 false misses of the dense retriever rests only on rejected passages (\S\ref{sec:rq4}).

\paragraph{What a false miss contains.}
For each false miss we classify the best evidence document named by the judge by its relation to the turn's gold. On both strong retrievers, 86\% of false misses are tied to the dialogue. On Fusion, 41\% rest on a \emph{same-thread negative}, a document mined as a hard negative for the same question, 14\% on a near-duplicate of a gold document (word 8-gram containment ${\geq}.8$), 19\% on the gold of another turn and 13\% on another chunk of the gold page. In the StackExchange domains, where each dialogue has its own pool of mined negatives, such negatives carry the answer in 63--69\% of false misses; in the BRIGHT domains the evidence is more often another turn's gold or a neighbouring chunk. The blind audit's own false misses fall into the same classes (90\% and 100\% tied to the dialogue). Dialogue-tied documents make up 72--74\% of all documents in \qnone{} contexts, so this shows where false misses sit rather than why; discounting every dialogue-tied false miss, precision still falls from .89 to .45.

\subsection{RQ3: The label changes conclusions about abstention}
\label{sec:rq3}

\begin{figure*}[t]
\centering
\includegraphics[width=\textwidth]{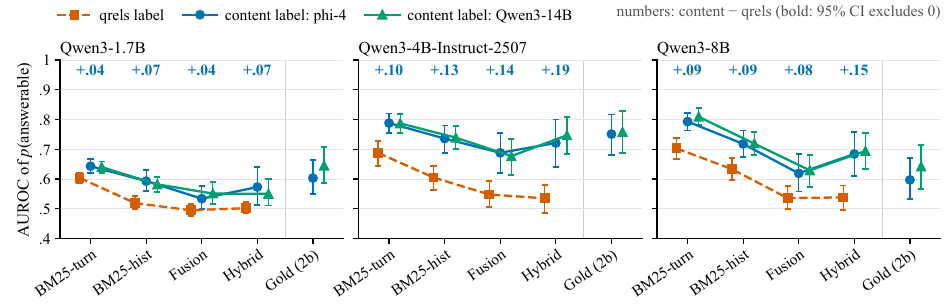}
\caption{AUROC of each reader's $p(\textsc{answerable})$ for separating sufficient from insufficient contexts, under the qrels label (\textsc{all} vs.\ \qnone) and the two content labels (\cyes{} vs.\ \cno). Gold contexts are all sufficient under the qrels label. Numbers: content (phi-4) minus qrels, paired over bootstrap resamples; bold if the 95\% CI excludes zero.}
\label{fig:auroc}
\end{figure*}

\begin{sloppypar}
\paragraph{Detector quality is understated.}
Under the phi-4 label, the AUROC of $p(\textsc{answerable})$ is significantly higher than under the qrels label for every reader at every retriever (Fig.~\ref{fig:auroc}). For Qwen3-4B the gap widens with the retriever (+.10, +.13, +.14, +.19). On the hybrid the qrels label puts all three readers near chance (.50--.54), while under the content label Qwen3-4B reaches .72 and Qwen3-8B .68. Restricted to contexts that both labels keep, the Qwen3-4B gap is still +.08 to +.16, so most of it comes from the label rather than from which contexts are dropped.
\end{sloppypar}

\paragraph{The qrels label hides the 2b/2c contrast.}
The question behind the 2b/2c pairing is whether a reader refuses more when its retrieval failed than when it had the gold. Under the qrels label this contrast is indistinguishable from zero already on BM25-hist for Qwen3-1.7B and on Fusion for Qwen3-4B. The content label keeps it significant for Qwen3-1.7B on BM25-hist and for Qwen3-4B on Fusion (+.20; not significant with refusals read from the Qwen3-8B answer judge) and the hybrid (+.31); for Qwen3-8B on Fusion the content contrast is not significant and the qrels contrast is significantly negative. Compared within turns, the content-label contrast stays significant on BM25-hist for all three readers, where the qrels contrast is not; on Fusion and the hybrid it is not significant for any reader (17--30 insufficient contexts per cell). The qrels contrast turns significantly negative for two readers on Fusion and all three on the hybrid, as if they refused less after a retrieval miss than with the gold of the same turn.

\paragraph{The wrong detector wins.}
Under the qrels label, a retrieval score that never reads the documents (the top-1 score, or the top-1 margin on Fusion) is the best insufficiency detector at every retriever: on BM25-hist the top-1 score reaches .73 (.71 when z-normalised within domain) against .63 for the best reader. Under the content label, the reader's own $p(\textsc{answerable})$ is best at every retriever (Qwen3-4B .74 on BM25-hist; on Fusion and the hybrid its lead is not significant), and the difference between the domain-normalised top-1 score and Qwen3-4B changes sign on BM25-turn (+.07 to $-.07$) and BM25-hist (+.11 to $-.17$), with all four intervals excluding zero. This is expected: the qrels label records whether the retriever found the gold, and the retriever's own confidence predicts that best.

\paragraph{How many conclusions change.}
We enumerate the comparisons an end-to-end study would report: refusal rates and AUROC between retrievers and between readers; refusal gaps between insufficient and sufficient contexts and between retrieved insufficient contexts and the gold (the 2b/2c contrast); AUROC against chance; and, for each change of retriever, the composition and rate terms of a Kitagawa decomposition of the refusal change~\cite{kitagawa1955components}. Of these 258 comparisons, 75 change sign or significance when the qrels label is replaced by the phi-4 label, 13 of them by sign; swapping one content judge for the other changes 25, none by sign. Of the 7 refusal-based changes that an answer judge can check, 4 survive with Qwen3-8B as answer judge and 6 with phi-4.

\subsection{RQ4: Replication on MTRAG}
\label{sec:rq4}

\begin{table*}[t]
\centering
\caption{MTRAG (777 answerable and partially answerable turns): the qrels label against the phi-4 content label for five first-stage retrievers, ordered by nDCG@10. Qs: BM25 over all user questions; Last: BM25 over the last turn; Rewr.: BM25 over MTRAG's rewritten question; Dense: Qwen3-Embedding-0.6B over the rewrite; Hybrid: equal-weight fusion of Rewr.\ and Dense. Bold: content-label AUROC whose difference from the qrels-label AUROC has a 95\% CI excluding zero.}
\label{tab:mtrag}
\setlength{\tabcolsep}{5pt}
\begin{tabular}{@{}lccccc@{}}
\toprule
 & Qs & Last & Rewr. & Hybrid & Dense \\
\midrule
nDCG@10 & .18 & .22 & .25 & .40 & .42 \\
no gold (\%) & 71 & 64 & 61 & 40 & 37 \\
content \cno{} (\%) & 46 & 42 & 37 & 20 & 18 \\
false misses (\%) & 23 & 23 & 27 & 40 & 38 \\
precision of ``no gold'' & .72 & .72 & .67 & .50 & .50 \\
\addlinespace[2pt]
Qwen3-4B AUROC: qrels & .73 & .76 & .70 & .67 & .68 \\
\quad content & .74 & .80 & \textbf{.78} & \textbf{.77} & \textbf{.75} \\
\bottomrule
\end{tabular}
\end{table*}

MTRAG~\cite{katsis2025mtrag} builds its qrels differently. Annotators held each conversation with a live RAG agent backed by an ELSER retriever, checked the passages it retrieved and could search for others; the passages they marked relevant are the qrels. There are no mined negatives, and the corpora are split into overlapping 512-token chunks. We use five first stages on the 777 turns that have qrels; with no training split, all settings were fixed in advance. phi-4 labels every top-5 context as on RECOR.

The decline replicates on a shorter ladder (Table~\ref{tab:mtrag}): precision falls from .72 to .50 (dense minus BM25 over all questions: $-.22$\ci{$-$.29}{$-$.16}), and false misses rise from 23\% to 38--40\%. Of the dense retriever's 109 false misses, 29 cite the gold of another turn of the same conversation, 21 another chunk of a gold passage's source document, and 56 only passages that the annotators never saw for that turn. The consequences replicate as well: Qwen3-4B's answerability AUROC is understated by the qrels label on the three strongest retrievers (+.07 to +.10), and on the hybrid its within-turn 2b/2c contrast is significant under the content label but not under the qrels label (+.22\ci{.06}{.38} on 58 turns against +.01\ci{$-$.06}{.09}; Qwen3-4B gives no status line on 13--23\% of MTRAG contexts, and these are excluded as on RECOR).

\section{Discussion and Recommendations}
\label{sec:discussion}

\paragraph{The protocol can give wrong conclusions.}
The reproduction overturns the qrels-derived sufficiency label. With a weak retriever most no-gold contexts are indeed insufficient (precision .68 on BM25-turn), but already on the official baseline 24--45\% of its \qnone{} contexts contain the answer, depending on the labeller (precision .44 under phi-4, .62 under the audit, .69 under Qwen3-14B). End-to-end benchmarks exist to compare systems with different retrievers, and that is exactly where the label fails. Readers that answer false misses correctly are penalised, and a detector that only tracks the retriever's confidence wins. The failure is not specific to RECOR's mined negatives: on MTRAG, with human qrels and no mined negatives, 38--40\% of the strong retrievers' ``insufficient'' contexts are sufficient (precision .50). We expect the cause to be general: qrels are collected per turn and per chunk around one retrieval setup, while the answer is spread over duplicates, sibling chunks and other turns' gold.

\paragraph{Recommendations.}
(1)~Do not compare abstention across retrievers, or select sufficiency detectors, under a qrels-derived label; if it must be used, report its precision on an audited sample for each retriever. (2)~Label sufficiency by content for each retrieved context, with judges from more than one model family, and validate each judge with a negative control first: small open models reproduce the trend but are too lenient for absolute rates. Human judgments, where they exist, should check the judge (\S\ref{sec:rq2}). (3)~Before using mined negatives as negatives, check them for the answer, not only for duplication against gold, and re-judge the gold of other turns and sibling chunks of gold pages as candidates. (4)~Report abstention conditionally on sufficiency and state the retriever. For RETECO 2c and similar tasks, this means scoring refusals against a content label of each submitted context rather than against the presence of gold.

\paragraph{Limitations.}
Sufficiency on RECOR is labelled by LLMs; the human evidence comes from MTRAG, and the judge-free copy count covers only part of the false misses. Item-level agreement is fair to moderate between the large judges and the audit ($\kappa$ .38--.49) and only fair for the small judges, so we rely on stage-level trends. All labellers see the reference answer, so a shared leniency toward on-topic documents would inflate the content label on strong retrievers; the negative control, the generation check and MTRAG's rejected passages bound this, but passages that no annotator saw, where most false misses lie, have no human check. The query expansions of Fusion and the hybrid were generated by Claude Sonnet~5, of the same family as the auditor. The readers are three Qwen3 models of 1.7B--8B parameters, and RECOR dev has 211 dialogues. We report many comparisons without correction; the counts of changed conclusions indicate scale, not significance.

\section{Conclusion}

We reproduced the qrels label for scoring abstention on RECOR, starting from an exact re-run of its official baselines. RQ1: no; the precision of ``no gold'' falls from .68 to .10. RQ2: labellers from five model families agree on the decline, and Qwen3-4B answers correctly from false misses. RQ3: the label understates answerability AUROC, hides the 2b/2c contrast and favours retrieval-score detectors. RQ4: the decline replicates on MTRAG.

\bibliography{custom}

\end{document}